\documentclass[%
 reprint,
 nofootinbib,superscriptaddress,twocolumn 
 amsmath,amssymb,
 prl,
]{revtex4-2}

\usepackage{graphicx}% Include figure files
\usepackage{dcolumn}% Align table columns on decimal point
\usepackage{bm}% bold math
\usepackage{amsmath}
\usepackage{wrapfig}
\usepackage{graphicx}
\usepackage{float}
\usepackage{subfigure}
\usepackage{amssymb}
\usepackage{bbold}
\usepackage{color}
\usepackage[normalem]{ulem}
\usepackage{comment}

\begin{document}

\preprint{APS/123-QED}

\title{ \hspace{1.1cm} Hybrid quantum image classification and federated learning \newline for hepatic steatosis diagnosis}

 \author{Luca Lusnig}
 \affiliation{Terra Quantum AG, 9000 St. Gallen, Switzerland}
 \affiliation{Research Unit of Paleoradiology and Allied Sciences, LTS - SCIT, Azienda Sanitaria Universitaria Giuliana Isontina, 34149 Trieste, Italy}

  \author{Asel Sagingalieva}
 \affiliation{Terra Quantum AG, 9000 St. Gallen, Switzerland}

 \author{Mikhail Surmach}
 \affiliation{Terra Quantum AG, 9000 St. Gallen, Switzerland}

 \author{Tatjana Protasevich}
 \affiliation{Terra Quantum AG, 9000 St. Gallen, Switzerland}

 \author{Ovidiu Michiu}
 \affiliation{Terra Quantum AG, 9000 St. Gallen, Switzerland}

 \author{Joseph McLoughlin}
 \affiliation{Terra Quantum AG, 9000 St. Gallen, Switzerland}

 \author{Christopher Mansell}
 \affiliation{Terra Quantum AG, 9000 St. Gallen, Switzerland}

 \author{Graziano de' Petris}
 \affiliation{Data Protection Officer, Azienda Sanitaria Universitaria Giuliana Isontina, 34149 Trieste, Italy}

 \author{Deborah Bonazza}
 \affiliation{Department of Medical, Surgical and Health Sciences, University of Trieste, Cattinara Academic Hospital, Trieste 34149, Italy}

 \author{Fabrizio Zanconati}
 \affiliation{Department of Medical, Surgical and Health Sciences, University of Trieste, Cattinara Academic Hospital, Trieste 34149, Italy}

 \author{Alexey Melnikov}
 \affiliation{Terra Quantum AG, 9000 St. Gallen, Switzerland}

 \author{Fabio Cavalli}
 \affiliation{Research Unit of Paleoradiology and Allied Sciences, LTS - SCIT, Azienda Sanitaria Universitaria Giuliana Isontina, 34149 Trieste, Italy}


\begin{abstract}
In the realm of liver transplantation, accurately determining hepatic steatosis levels is crucial. Recognizing the essential need for improved diagnostic precision, particularly for optimizing diagnosis time by swiftly handling easy-to-solve cases and allowing the expert time to focus on more complex cases, this study aims to develop cutting-edge algorithms that enhance the classification of liver biopsy images. Additionally, the challenge of maintaining data privacy arises when creating automated algorithmic solutions, as sharing patient data between hospitals is restricted, further complicating the development and validation process. This research tackles diagnostic accuracy by leveraging novel techniques from the rapidly evolving field of quantum machine learning, known for their superior generalization abilities. Concurrently, it addresses privacy concerns through the implementation of privacy-conscious collaborative machine learning with federated learning. We introduce a hybrid quantum neural network model that leverages real-world clinical data to assess non-alcoholic liver steatosis accurately. This model achieves an image classification accuracy of $97\%$, surpassing traditional methods by $1.8\%$. Moreover, by employing a federated learning approach that allows data from different clients to be shared while ensuring privacy, we maintain an accuracy rate exceeding $90\%$. This initiative marks a significant step towards a scalable, collaborative, efficient, and dependable computational framework that aids clinical pathologists in their daily diagnostic~tasks.
\end{abstract}

\maketitle

\section{Introduction}
\begin{center}
\fbox{
\begin{minipage}{0.45\textwidth}
Please check the published version, which includes all the latest additions and corrections: Diagnostics 14(5):558, 2024, DOI: \href{https://doi.org/10.3390/diagnostics14050558}{10.3390/diagnostics14050558}
\end{minipage}
}
\end{center}

In addressing the global challenge of determining liver viability for transplantation, this study tackles two main issues: the accuracy of hepatic steatosis diagnostics and the preservation of patient data privacy. The accurate classification of liver biopsy images is vital for assessing transplant viability, yet it is hampered by the substantial data requirements for training sophisticated machine learning models and the inherent privacy concerns associated with sensitive patient data. Federated learning (FL) offers a solution to these privacy concerns by enabling collaborative model training across multiple clients without centralizing sensitive data, thus adhering to data protection regulations such as the GDPR~\cite{GDPR2016a}. This approach is further underscored by upcoming regulations, like the EU AI Law~\cite{EU_AI_law}, emphasizing the need for models that balance accuracy with privacy.

Recent advancements in convolutional neural networks (CNNs)~\cite{LeCun1999, Rosenblatt1963PRINCIPLESON} and the extension to residual neural networks (ResNet)~\cite{ResNet} has  shown promise in various pattern recognition tasks, including medical image analysis. The evolution of machine learning has seen the rise of quantum machine learning (QML), which combines quantum computing principles with classical techniques to potentially enhance model performance~\cite{Schuld2018, kordzanganeh2023benchmarking, biamonte2017quantum, kordzanganeh2023parallel}. The integration of hybrid quantum--classical neural networks (HQNNs) into traditional CNN architectures promises improved accuracy, even with reduced dataset sizes, by leveraging the unique capabilities of quantum computations.

Our study specifically addresses the challenge of diagnosing hepatic steatosis by creating a dataset of 4400 samples evenly distributed across four steatosis stages, then categorized into transplantable and non-transplantable livers based on steatosis level. We aimed to surpass the diagnostic accuracy of human experts, targeting over $90\%$ accuracy with a false negative rate below $5\%$. Utilizing a ResNet and an HQNN, we achieved remarkable accuracy levels, demonstrating the effectiveness of these models even on reduced datasets.

The federated learning approach we employed allowed for the training of a classical CNN with data from multiple clients, up to 32, without compromising data privacy, thus adhering to GDPR standards. This method achieved comparable accuracy to centralized models while utilizing significantly fewer data samples per client.

This research not only presents a novel application of QML and federated learning to the field of liver biopsy image classification but also sets a precedent for addressing privacy and data scarcity issues in medical diagnostics. Our findings suggest that HQNNs, combined with a federated learning framework, could become the preferred model for future applications where data privacy and availability are of concern. We plan to extend this work by further refining the steatosis staging and exploring quantum federated learning algorithms, contributing to the broader application of QML in healthcare.

\section*{Literature Review}\label{LiteratureReview}

In this subsection, we provide a literature review that focuses on existing works where at least two of the main themes of our research intersect. 
As introduced above, these main themes are: %(1) 
the diagnosis of liver disease, specifically hepatic steatosis; %(2) 
privacy enhancements via federated learning; %(3) 
quantum technology and %(4) 
machine learning, especially for image classification tasks. 
Due to the large number of research papers in these domains, we mostly refer the reader to recent review articles. 
However, for each intersection of the domains, we highlight the important results and the current limitations.
A discussion of their promise and potential for progress is given in Section~\ref{future_works}.

\subsubsection{Federated Learning and Medical~Diagnosis} %(1-2)
%%%%%%%%%%Important results
FL has been employed in hospitals and universities in several countries, often in collaboration with companies like Intel or Nvidia (see Section~6 of Ref.~\cite{FLinSmartHealthCare} and references therein).
%%%%%%%%%%Current limitations
Despite the success of these real-world projects, FL is still an emerging discipline.
As such, there are several challenges that require further research, including standardization, efficiency, economic viability, and data quality. 

\subsubsection{Artificial Intelligence and Liver~Disease} %(1-4)
There are several recent reviews on the use of artificial intelligence in liver\linebreak  \mbox{diseases~\cite{nishida2023artificial, nam2022artificial, allaume2023artificial}}.
 %%%%%%%%%%Important results
It is now well-established that modern machine learning methods can perform extremely well in image classification tasks. 
Since images of liver tissue are central to the diagnosis of liver pathologies, this is a natural application area for machine learning models and indeed, the state-of-the-art results are comparable to the performance of experts in the field.
%%%%%%%%%%Current limitations
However, large-scale clinical trials with external validation cohorts still need to be performed.

\subsubsection{Quantum Security And~Privacy} %(2-3) 
%%%%%%%%%%Important results
Can information be processed and transmitted in a secure fashion? The field of quantum cryptography provides an affirmative answer to this question. However, instead of relying on computational assumptions as in classical cryptography, the security and privacy of the information is ensured by the laws of quantum mechanics~\cite{Security2022Portman}. While considerable attention has been given to the development of quantum key distribution systems, various other quantum communication protocols exist, each of which has a number of advantages. A relevant example of distributed computing is the comparison between classical homomorphic encryption and blind quantum computation. The former protocol employs encryption to stop the server from learning anything about the inputs or outputs of the algorithm that it runs on behalf of the client. The latter protocol can not only do this but can also prevent the server from finding out what the algorithm is~\cite{Blind2013Morimae}. Further examples include quantum distributed learning~\cite{gilboa2023exponential} and quantum federated learning~\cite{Ren2023TowardsQFL}. 

%%%%%%%%%%Current limitations
Currently, many quantum communication protocols have been implemented in laboratory settings but few have been demonstrated in more practical scenarios. The challenges surrounding this largely relate to the technological immaturity of some important hardware components~\cite{Pettit_2023, Azuma_2023}. 

\subsubsection{Quantum Computation and Machine~Learning} %(3-4) 
%%%%%%%%%%Important results
QML is a burgeoning field at the intersection of quantum mechanics and classical machine learning~\cite{cerezo2022challenges}. 
It leverages uniquely quantum phenomena, such as superposition and entanglement, and has the potential to offer exponential speedups in specific computational tasks~\cite{Schuld_2014, biamonte2017quantum, Dunjko.Briegel.2018, qml_review_2023}. 
By operating within an exponentially large search space, these algorithms can achieve enhanced efficiency in pattern recognition and prediction~\cite{Lloyd2013, 365700, Lloyd1996UniversalQS, schuld2020circuit}.
Recent advancements in quantum and quantum-inspired computing have shown further possibilities for refining and enhancing classical machine learning techniques~\cite{Neven2012QBoostLS, PhysRevLett.113.130503, 
saggio2021experimental, rainjonneau2023quantum, riaz2023accurate, senokosov2024quantum, naumov2023tetra}. 
In particular, HQNNs have garnered attention for their applications across industries~\cite{sedykh2023quantum, haboury2023supervised}. Further, HQNNs have exhibited faster convergence to optimal solutions and superior accuracy, especially with small datasets, compared to their classical counterparts~\cite{pharma, asel1}. 

%%%%%%%%%%Current limitations
Despite its immense potential, QML is not without challenges. The sensitive nature of quantum computations demands rigorous error correction and makes them vulnerable to external disturbances~\cite{Aaronson2015}.
The main challenge that quantum neural networks face today is the noise-free barren plateau problem~\cite{bp}. This issue can make the optimization landscape almost flat, making it very hard for optimization algorithms to make meaningful updates to the parameters during training. Additionally, when using a neural network to handle large datasets, one needs a large number of qubits, and this exceeds the capabilities of the fault-tolerant quantum simulators and hardware available today. 

\subsubsection{Quantum Machine Learning and~Medicine} %(1-3)
%%%%%%%%%%Important results
In Ref.~\cite{ajlouni2023medical}, a hybrid quantum convolutional neural network was used in a brain tumor classification task.
The accuracy of this model converged more quickly than a conventional CNN (70 epochs rather than 100). 
Other applications of HQNNs in healthcare and pharmaceuticals include Refs.~\cite{pharma, McArdle2020QuantumCC, Nannicini, fedorov, moussa2023application}.
%%%%%%%%%%Current limitations
Unfortunately, due to the limited availability of quantum computers and the difficulty of simulating their information processing capabilities using a classical computer, studies of QML in medicine are currently rather small in number.

\section*{Materials and methods}\label{methods}

This study focuses on non-alcoholic liver steatosis, which is both common and on the rise in many countries.
In Section~\ref{NAFLD}, we describe several aspects of this disease and explain how its rapid and adequate diagnosis can inform decisions about the suitability of liver transplantations.
In order to realize this project, we obtained data from the anonymous teaching archive of the Institute of Pathological Anatomy of the University Clinical Department of Medical Surgical and Health Sciences, University of Trieste. 
We provide details of these data in Section~\ref{dataset}.
Our two main methods, QML and FL, are described in Section~\ref{QML} and Section~\ref{Federated_learning}, respectively. 

\subsection{Non-Alcoholic Fatty Liver Disease (NAFLD)}\label{NAFLD}

Steatosis can be defined as an abnormal accumulation of fat within the liver cell and can occur in various liver diseases and its clinical implications generally correlate with its extent. Non-alcoholic fatty liver disease (NAFLD) is a chronic disorder of relatively recent identification but in considerable and continuous epidemiological growth. The afflicted population is estimated to be about 1 billion people worldwide~\cite{Castera2019}. NAFLD ranges from simple steatosis in the absence of excessive alcohol intake to non-alcoholic steatohepatitis (NASH) with or without cirrhosis~\cite{dhamija2019non}. 

The prevalence of non-alcoholic hepatic steatosis is steadily increasing to about 25\% in the general population (compared to 1.5--6.45\% for NASH), favouring obese and/or diabetic patients or, more generally, those with metabolic syndrome~\cite{diehl2017cause,loria2010practice,taneja2020nonalcoholic,araujo2018global}.
As for the incidence, it was estimated to be about two new cases/100 people/year~\cite{loria2010practice,ratziu2010position}.
In addition to genetic and non-modifiable factors (age and sex), environmental and ethno-geographical factors must also be considered, remapping the purely multifactorial etiology of this condition, but also how it is related not only to the quantity and quality of food intake, which is why there is an increase in prevalence, around 8.7--20, even in many developing countries~\cite{zhou2019noninvasive,benedict2017non,ma2020proportion}.

The natural history of NAFLD progresses from steatosis, steatohepatitis, fibrosis, cirrhosis and hepatocellular carcinoma~\cite{dhamija2019non}. Approximately 30\% of the individuals with NAFLD show histologic progression: 15–20\% of these individuals may develop cirrhosis, while another 30–40\% may suffer liver-related morbidity and mortality~\cite{idilman2013hepatic}. Currently, cirrhosis due to NASH is the second leading indication for liver transplantation in the United States~\cite{wong2015nonalcoholic}. 

Hepatic steatosis is a pathogenic, potentially reversible condition, even in serious cases. Furthermore, the presence of steatosis has emerged as the single most important variable determining graft function after transplantation~\cite{nocito2006steatosis} and can be considered one of the key parameters in the decision on the suitability of a liver transplant~\cite{chu2015donor, cesaretti2019assessment}.  It is, therefore, important to identify the condition and assess its severity~\cite{reeder2010quantification}.

\begin{figure*}[ht!]
    \centering
    \includegraphics[width=1\linewidth]{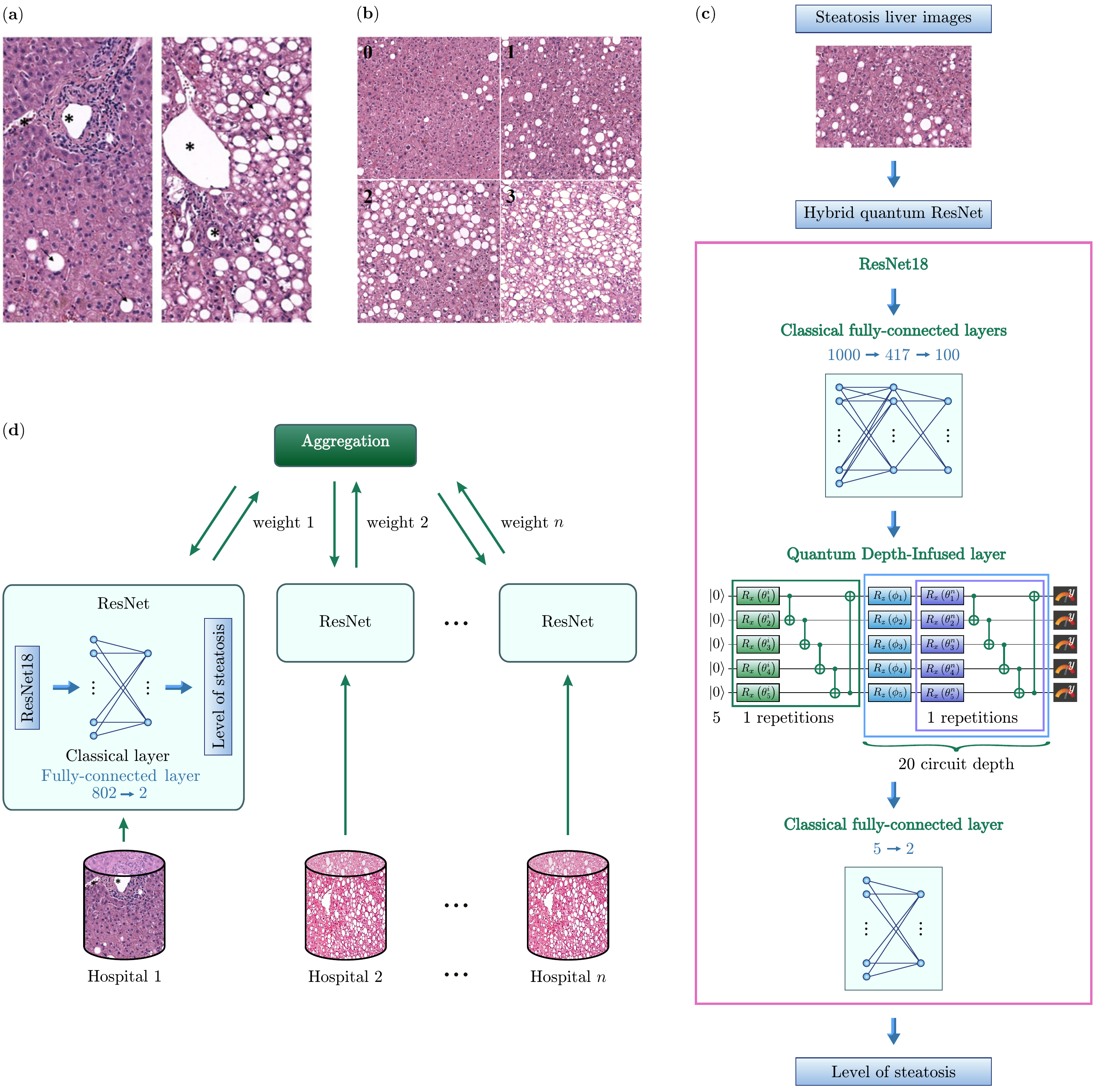}
    \caption{(\textbf{a}) Hepatic tissue with steatosis. %DIF > FIGURE ARE OK.
The~image on the right shows a hepatic tissue with a severe degree of steatosis. Fat droplets are marked with black arrow. Vessels are marked with asterisk. (\textbf{b}) 0: liver biopsy with score 0; 1: in this field, hepatocytes have steatosis between 5 and $33\%$ (score 1); 2: liver biopsy with macrovesicular steatosis between 33 and $66\%$ with an inhomogeneous distribution of fat drops (score 2); 3: steatosis over $66\%$ (score 3). Images of \mbox{[1024 $\times$ 1024]} pixel. Hematoxylin and eosin, 20$\times$. (\textbf{c}) Architecture of a Hybrid quantum ResNet model for analyzing liver biopsy images. It starts with a classical ResNet18 neural network, followed by two fully connected layers. The~information is then relayed to a QDI layer consisting of $5$ qubits and $20$~variational layers. After~quantum processing, a~classical 5-element vector is inputted into another fully connected layer that determines the suitability of the liver for transplantation. (\textbf{d})~Horizontal federated learning architecture for analyzing liver biopsy images. The~cylinders represent the datasets that are used to train a classical ResNet18, both of which are physically present in the individual hospitals. The~weights of the ResNet are then sent to a server (i.e., aggregation) which processes them and returns the updated weights to the individual~ResNet18.}
    \label{fig:HQNN}
\end{figure*}

Histologically, steatosis is evaluated in paraffin-embedded sections stained with hematoxylin and eosin 
%(HE) 
and is visible as white droplets in the cytoplasm of hepatocytes. The fat dissolves during histological processing and droplets appear as gaps in the tissue. Droplets of fat can be distinguished from other empty spaces, such as vessels or tissue artifacts, by their size and rounded shape (Figure~\ref{fig:HQNN}a).

Liver biopsy and histological classification are the gold standard for diagnosing hepatic steatosis. It also allows for the identification of other histological features such as steatosis zonality, droplet size (macrovesicular vs. microvesicular), inflammation, cellular injury, iron overload and fibrosis staging. 

Kleiner~et~al.~\cite{kleiner2005design} proposed a histological scoring system for NAFLD, the NAFLD Activity Score (NAS), to include a wider spectrum of NAFLD that includes isolated steatosis and not only steatohepatitis. The authors analyzed the degree of inter‐observer and intra‐observer agreement between and within pathologists. The semi-quantitative evaluation of steatosis shows an inter-rater k value of $0.79$, while the semi-quantitative evaluation of steatosis intra‐rater agreement is $0.83$ (with $k = 1$ for perfect agreement and values of k near 0 for random agreement).

In Brunt~et~al. classification and in Kleiner~et~al. classification, steatosis was classified from grade $0$ to grade $3$~\cite{brunt1999nonalcoholic, kleiner2005design}. The extent of steatosis is commonly evaluated and reported semi-quantitatively (Table~\ref{tab:my_label}).

\begin{table}[ht!]
    \centering
    \begin{tabular}{c|c}
\hline
Steatosis percentage& Steatosis grade \\
\hline
\textless5\%&0\\
5--33 \%&1\\
33--66\% &2 \\
\textgreater66  \%&3 \\
\hline
\end{tabular}
    \caption{Staging of steatosis according to the percentage of fat in the liver~\cite{brunt1999nonalcoholic, kleiner2005design}}
    \label{tab:my_label}
\end{table}

Taking into account that the global prevalence of NAFLD is between 24 and 25\% with some regional differences (the highest rates are reported in South America and the Middle East, followed by Asia, USA and Europe)~\cite{araujo2018global}, it is not surprising that steatosis represents the most frequent liver morphological alteration in the population of potential transplantable liver donors. A model such as the one proposed, even in its current formulation could, therefore, be considered a rapid diagnostic system that can be used in a very important decision-making step in liver transplantation.

\subsection{Dataset}\label{dataset}

Histological bioptic liver specimens were extracted from the anonymous teaching archive of the Institute of Pathological Anatomy of the University Clinical Department of Medical Surgical and Health Sciences, University of Trieste. 
The liver biopsy was fixed in $10\%$ formalin, embedded in paraffin and stained with hematoxylin-eosin.
Specimens already classified as NAFLD were rechecked and restaged by two experienced pathologists (FZ and DB) before being part of the dataset.

The dataset consists of $41$ jpeg2000 images of liver tissue digitized by the D-Sight 2.0 microscope manufactured by Menarini\textsuperscript{®} s.r.l. of Florence (Italy). %MDPI: Please state the name of the manufacturer, city, and country from where the equipment was sourced LUSNIG: DONE.
 Each of the jpeg2000 images represents a patient and has a size varying between 80,000 and 120,000 pixels per~side. 

In order to have datasets with a significant number of images of equal size, each jpeg2000 image was divided into smaller images of [1024 $\times$ 1024] pixels (Figure~\ref{fig:HQNN}b). Henceforth, we will refer to these portions of the image as~samples.

The subdivision of the original image into a set of smaller images is justified by the fact that the physician during diagnosis visually analyzes the image by unconsciously dividing it into sectors, because there may be areas where the disease is more or less extensive, thus representing a different staging.

We created a balanced dataset containing $1100$ images for each stage of the disease, then in the case study, we divided the images in the dataset into two classes, the one representing livers suitable (i.e., samples labeled $0$ and $1$) for transplantation and those not suitable (i.e., samples labeled $2$ and $3$) while maintaining a balanced dataset.

\subsection{Quantum machine learning}\label{QML}

In this study, we have to employ such a quantum layer to be able to propagate $100$ classical output neurons. Given the challenges mentioned in Section~\ref{LiteratureReview}, it is advisable to use quantum layers with fewer qubits that can still handle large datasets. This approach will also reduce the noise related to the physical implementation of qubits, which will help with the barren plateau problem, allowing us to potentially run our neural network on a physical quantum processor in the future.

Thus, we include the Quantum Depth-Infused (QDI) layer~\cite{pharma} in our work~\mbox{(Figure \ref{fig:HQNN}c)}. In~ this layer, each of the $100$ features before the quantum layer is encoded on a lattice (blue rectangles with $R_Z$ sign in Figure~\ref{fig:HQNN}c). The~first five features are on the first lattice length, features 6--10 on the second length, and so on. Using the ''angle embedding'' method, we encode these classical features into the quantum Hilbert space. This is achieved by rotating each qubit in the ground state around the $Z$-axis on the Bloch sphere by an angle proportional to the corresponding value in the input vector. To make the quantum layer part of the neural network and adapt it to the data, we use variational layers (purple rectangle in Figure~\ref{fig:HQNN}c). In~ QDI, each variational layer has two parts: rotations with trainable parameters (in our case rotation around the $X$-axis) and subsequent CNOT operations. The rotations act as quantum gates, changing the encoded input data based on variational parameters $\theta_i^j$. The CNOT operations entangle the qubits in the quantum layer, creating quantum superposition. During the measurement phase, all qubits, except the first, undergo a CNOT operation on the first qubit, spreading the $Y$-measurement to all qubits and transforming the information from the Hilbert space vector to a classical five-neuron~output.

Given its outstanding performance in a distinct medical task~\cite{pharma}, it was hypothesized that this same layer would effectively broaden the function space described by the proposed neural network. At the same time, it can efficiently encode a significant amount of information in a limited number of qubits. Therefore, this layer potentially can help to avoid the problems associated with quantum computers now. These advantages became the primary reasons why we chose this architecture for a quantum layer.

%These innovations emphasize the potential of HQNNs over purely classical methods. With this arsenal of quantum tools, we are better equipped to address intricate medical challenges, such as the task of steatosis liver image classification. Such quantum-enhanced methodologies may revolutionize diagnostics and therapeutic strategies in the near future~\cite{Cordier_2022}.

These innovations highlight the potential of HQNN compared to purely classical methods. We believe that improvements in the efficiency and accuracy of machine learning models by integrating quantum layers into a classical framework can only be definitively verified empirically. Therefore, we conducted experiments on a static dataset including liver images. Empirical data obtained from these experiments demonstrate an improvement in classification performance when using HQNN compared to classical counterparts. Such quantum-enhanced methodologies have the potential to revolutionize diagnostics and therapeutic strategies in the near future~\cite{Cordier_2022}.

\subsection{Federated~Learning}\label{Federated_learning}

Federated learning~\cite{pmlr-v54-mcmahan17a} is a machine learning technique to achieve privacy-preserving collaborative learning among different parties. Compared with traditional centralized and distributed learning, which require data to be collected in a single compute node in the former and in a server to parallelize over multiple nodes the computation and thus speed up the training phase in the latter, in federated learning only locally trained models are exchanged, without sharing the data with the consequence of guaranteeing privacy.

In addition, we can improve data security and privacy by using privacy-enhancing technologies~\cite{kaaniche2020privacy}, which are developed to meet the basic principles of data protection, minimizing the use of personal data and maximizing data security. The two most widely used techniques are Fully Homomorphic Encryption (FHE)~\cite{rivest1978data} and Secure Multi-Party computation (SMPC)~\cite{ZHAO2019357}. 

FHE is a key-based encryption in which the encryption process preserves algebraic operations, in this technique a third party can calculate the data without knowing it. In addition to depending on a trusted third party, they do not provide perfect security, and therefore, could represent a lack of security when used in FL. To solve this problem, Liang et al. have proposed a symmetric quantum version of FHE~\cite{liang2013symmetric} that guarantees perfect privacy by making such an encryption technique safe.

The SMPC technique is a cryptographic protocol that distributes computation across multiple parties where no individual party can see the other party's data. This approach does not need a trusted third party to see the data, and in addition, the data are encrypted in-use because they are divided and distributed among the players during calculation, which makes it secure against quantum attacks (for now). On the other hand, there is a high communication cost between players because of the computational overhead due to the use of random numbers which must be generated in order to ensure the security of the computation.

% \begin{quote}
% This is an example of a quote.
% \end{quote}

%%%%%%%%%%%%%%%%%%%%%%%%%%%%%%%%%%%%%%%%%%

\section*{Results}\label{material_&_methods}

In Section~\ref{advantages}, we present the image classification results for our HQNN and the classical ResNet model.
In Section~\ref{m&m_FL}, we provide the results related to the FL setting, where each client accesses only a small proportion of the total dataset. 

\subsection{Hybrid Quantum Image~Classification}\label{advantages}

We address the classification of liver steatosis types by harnessing the capabilities of transfer learning. Transfer learning, a potent strategy in neural network training, allows expertise gained from one problem to bolster performance on another  related problem~\cite{Transfer_Learning}.

For our classification task, we deploy the ResNet model, pretrained on the expansive ImageNet dataset~\cite{ImageNet, ResNet}. Training deep neural networks is often plagued by the vanishing gradient issue, but ResNet effectively navigates this with its innovative residual blocks. These blocks enable deeper layers to retain essential input data information by directly passing inputs to subsequent layers within the block. Owing to this design, ResNet has cemented its reputation as a formidable tool for image classification tasks. Specifically, we employ ResNet18 via PyTorch~\cite{PyTorch}, with ``18'' indicating the network's depth in terms of layers. As~illustrated in Figure~\ref{fig:HQNN}c, subsequent to the ResNet architecture, we add a fully-connected layer. Each neuron in this layer aligns with a distinct class in our binary classification objective.

Our proposed hybrid quantum model called hybrid quantum ResNet is depicted in Figure~\ref{fig:HQNN}c. Upon~ the completion of processing through ResNet, the hybrid architecture includes both classical and quantum layers. Given that the output feature map from ResNet encompasses $1000$ distinct features, the subsequent fully-connected layers incorporate these $1000$ input neurons, producing $100$ output features.  We used the Pennylane 0.29.0 ~\citep{ber_pen_2022} software development kit for implementing HQNNs, given its efficient integration with PyTorch. 

After classical processing, QDI seamlessly translates the output features into the quantum domain. Crucially, the QDI structure adeptly manages $100$ features in our model using a minimalistic setup of just $5$ qubits, achieved through the synergy of $20$ data re-uploading layers, each coupled with a variational layer. This architectural efficiency, where $5$ qubits handle $20$ data re-uploading layers, collectively corresponds to the processing of $100$ features. This streamlined approach is particularly valuable in the Noisy Intermediate-Scale Quantum (NISQ) era.

The QDI layer's architecture initiates with a series of variational layers, seamlessly integrating with $20$ data re-uploading layers, each complemented by a variational layer. Importantly, the subsequent paragraph emphasizes that the cumulative effect of these layers surpasses a total of $105$ variational gates. Following these processing steps, measurements are taken from all qubits, converting the quantum data into a classical vector. This vector is then fed into a final classical fully-connected layer, effectively discerning and predicting the corresponding output class.

The whole dataset contains $4400$ images and, to speed up the training process without losing the prediction quality of our model, we cropped the images to size $258 \times 258$. 

In this comprehensive investigation, hyperparameter optimization played a pivotal and intricate role in refining both classical and quantum-hybrid models. Leveraging the Optuna Python library~\cite{optuna_link}, we meticulously optimized various ResNet configurations for the classical model, including ResNet$18$, $34$, $50$, $101$, and $152$. The architectural parameters of linear layers post-ResNet, such as the number of layers, neurons in each layer, and activation functions between layers, along with hyperparameters of the optimizer (including optimizer type and specific parameters for each optimizer, such as learning rate, epsilon, and betas for Adam optimizer), underwent scrupulous optimization to ensure the attainment of acceptable model output (with $100$ neurons as the desired output). 

For the hybrid model, in addition to the previously mentioned aspects, we conducted a meticulous hyperparameter optimization process associated with the QDI layer. This involved optimizing the number of qubits, variational layers, types of variational gates, types of encoders, and measurement procedures. Preprocessing parameters for the dataset and dataloader, such as image cropping size and the number of images in each batch were also subject to optimization. This intricate optimization process unfolded iteratively, with the number of Optuna trials spanning from $50$ to $150$. The adjustments made to all parameters were precisely calibrated to achieve optimal accuracy for the resulting models.

Hybrid and classical neural networks were trained for $30$ epochs with Adam~\cite{Adam,kingma2014adam}. For our loss computation, we adopt the cross-entropy formula:
\begin{equation}
l = -\sum_{c=1}^{k}{y_c \log p_c}
\end{equation}
Here, $p_c$ symbolizes the predicted probability for a class, and $y_c$ is a binary value, $0$ or $1$, designating whether an image corresponds to the predicted class. The total class count is denoted by $k$. 

\begin{figure*}[ht!]
    \centering
    \includegraphics[width=1\linewidth]{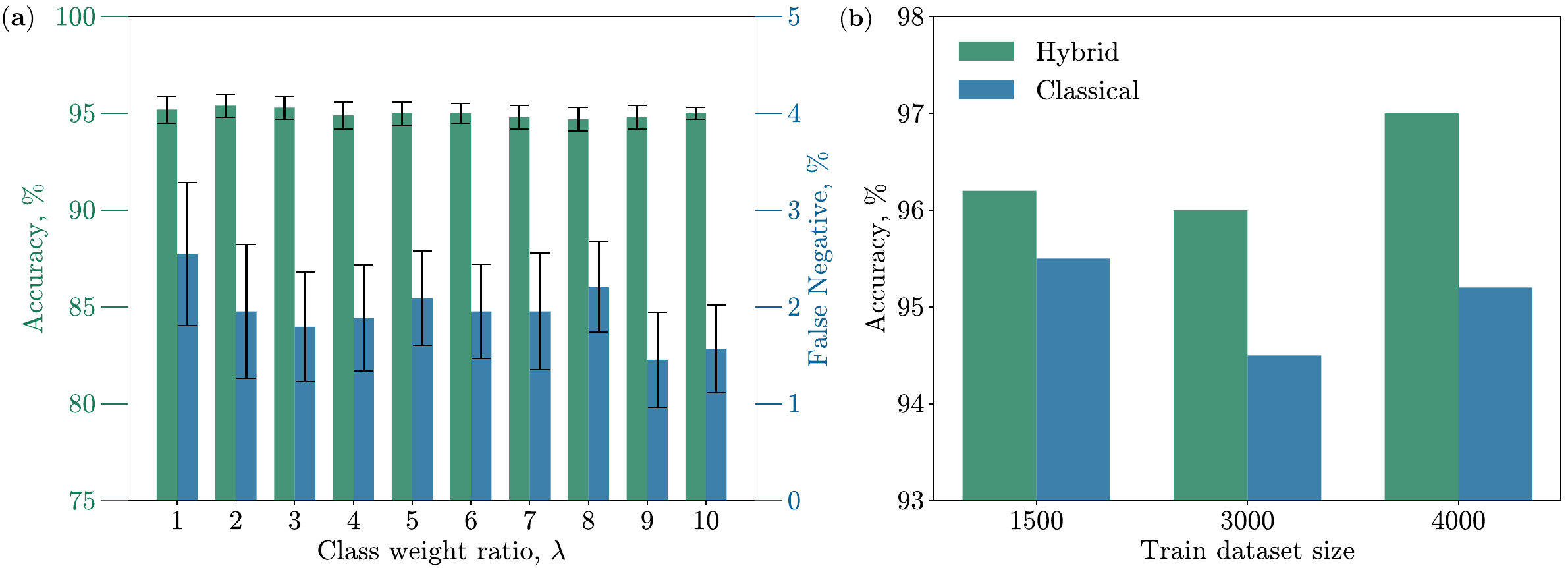}
    \caption{(\textbf{a}) The graph presents the correlation between classification accuracy (depicted by green bars) and the percentage of false negative answers in classification of hepatic steatosis stage (represented by blue bars) of the classical model, relative to the class weight ratio, $\lambda$. The~black bars indicate the standard deviation of values observed during model training using $5$-fold cross-validation. (\textbf{b})~The relationship between the classification accuracy of hepatic steatosis and the size of the training dataset for both the hybrid model (illustrated in green) and the classical model (depicted in blue) with $\lambda = 1$. The~testing set contains $400$ images for all experiments depicted in this figure.}
    \label{fig:result}
\end{figure*}

Figure~\ref{fig:result}b shows the result of the experiment; it can be seen that the hybrid model achieves an accuracy of $97\%$, while the classical one only reaches $95.2\%$. To confirm the results that the hybrid model performs well when there is not a large amount of data, we conducted an experiment, reducing the training set, and~showed in Figure~\ref{fig:result}b that even on $1500$ images, the quality of the hybrid network is not lower than $96\%$ and is always better than the accuracy of the classical neural network. Moreover, the number of weights in the hybrid model is $1.75$ times less than in the classical model.

Recognizing the critical importance of minimizing false negatives in our classification, we incorporate an additional regularization term during training. This ensures a heftier penalty for false negative predictions. The penalty multiplier for false negatives is represented by the parameter $\lambda$. In this case, the formula for cross-entropy loss with class weights looks as follows:
\begin{equation}
l_{\lambda} = -\sum_{c=1}^{k} w_c \cdot y_c \cdot \log(p_c)
\end{equation}
where $l_{\lambda}$ --- cross-entropy loss, $k$ --- total number of classes, $w_c$ --- weight for class $c$ ($1$ or $\lambda$), $y_c$ --- binary value (0 or 1) indicating whether an image corresponds to the predicted class, $p_c$ --- predicted probability for class $c$. This formula takes into account the class weights when computing the cross-entropy loss.

It is also worth noting that in all experiments, the model's false negative rate was below $5\%$. The dependence of the prediction accuracy of the classical ResNet on $\lambda$ is plotted in Figure~\ref{fig:result}a. It shows that the lower the percentage of false negative responses, the lower the overall prediction accuracy of the model, so it is important to find the tradeoff between the $\lambda$ and the overall percentage of accuracy. However, for any $\lambda$ from $1$ to $10$, the false negative rate is below $3\%$.

In Figure~\ref{fig:result}b, the~ classification accuracy is plotted against the size of the training dataset. It shows that the hybrid model achieves superior classification accuracy even when trained on less than $40\%$ of the dataset. Additionally, the percentage of false negative responses in all experiments remained below the $3\%$ threshold, indicating the reliability of the model's performance. Furthermore, the model exhibits robustness concerning $\lambda$, with a declining trend in the percentage of false-negative responses as $\lambda$ increases. It is noteworthy that an increase in $\lambda$ results in a rise in classification accuracy (accompanied by a reduction in false negatives) in certain cases, such as when $\lambda = 2$.

\subsection{Federated Learning for Medical~Privacy}\label{m&m_FL}
In this section, we simulate in detail a real-world scenario in which a variable number of hospitals, called clients, want to create a common model based on the data present locally in each individual hospital without having to physically share the data with each other or with an external server.

Let us assume that the hospitals involved in the simulations use the same type of data and a common model has been proposed and shared.
By subdividing the datasets wisely and avoiding the same data being used multiple times in datasets belonging to different hospitals, we can effectively simulate a federated approach while guaranteeing data privacy and eliminating the sharing of sensitive data by only exchanging between individual hospitals and aggregation servers, the parameters of individual local models.

\begin{figure}[ht]\label{FL}
    \includegraphics[width=1\linewidth]{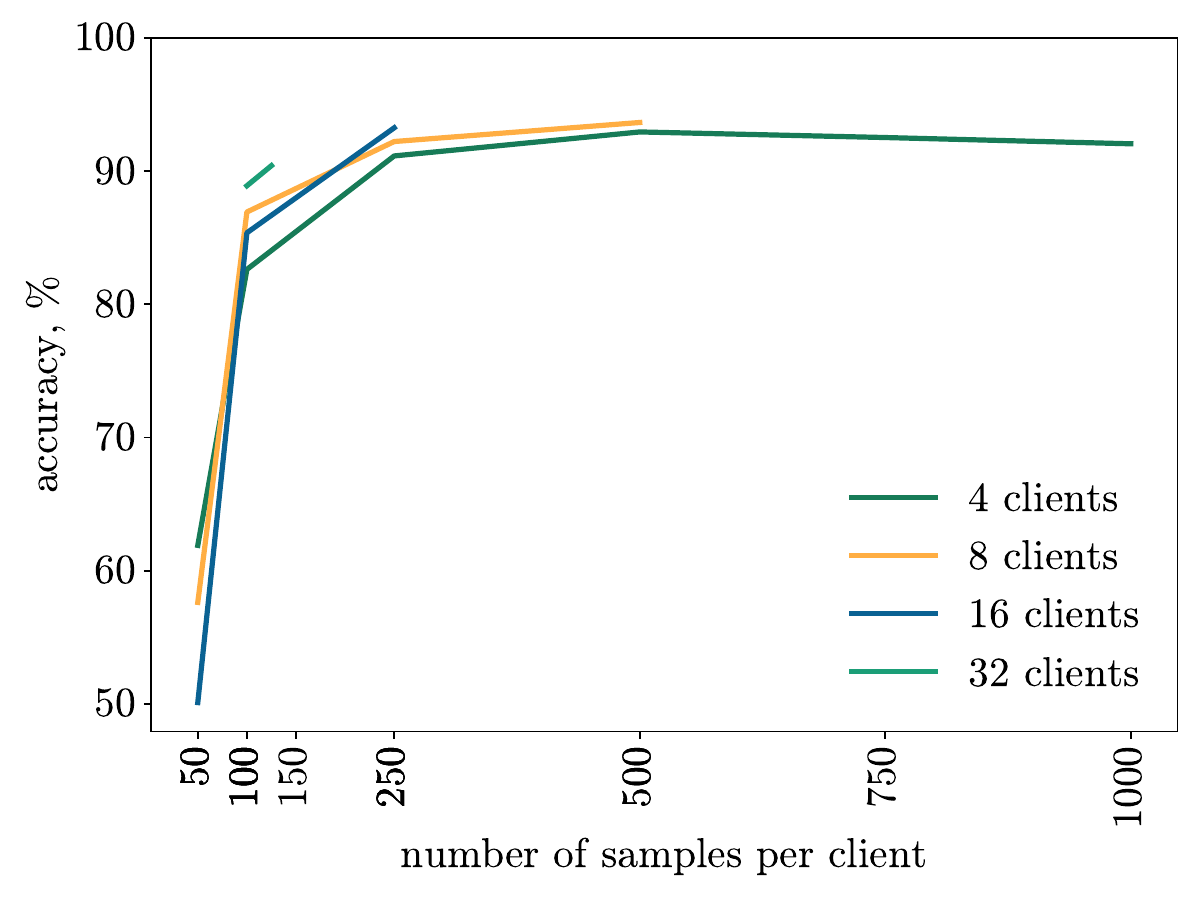}
    \caption{
    Behavior of different architectures as the number of samples available in the dataset changes.
    }
    \label{fig:acc_combined}
\end{figure}

The results in Figure ~\ref{fig:acc_combined} show that as the number of hospitals involved increases, the number of samples to be collected and labeled in each individual hospital decreases significantly, keeping the model performing well from an accuracy point of view.
This situation also makes this approach a viable alternative to centralized systems from a feasibility point of view, since it is not always possible for a single hospital to collect the amount of data needed to train a performant~model.

 Hereafter, we describe in detail the architecture and strategy as well as the dataset used for the federated learning solution.

For implementing the federated learning solution, we use the Flower framework~\cite{beutel2020flower} which is open source and provides support for extending FL implementations to mobile and wireless clients, with heterogeneous compute, memory, and network resources. In~addition, it is released under an Apache 2.0 license, making it a perfect candidate for integrating a QML library.

The proposed architecture is of the horizontal federated learning type, in which each client trains its own model on its own device and then sends its parameters to the server, which after aggregating them sends a new model to the clients. The aggregation strategy used is the vanilla federated average~\cite{pmlr-v54-mcmahan17a} strategy, in which all clients were used in both the training and validation phases.

The data set was balanced where the number of samples available for each client was the same and equally divided between the two classes, and no samples were used by two different clients.

In all simulations, the common model to be trained is the same as in the centralized case and in order to avoid overfitting in the training phase in each client, the number of training epochs was less than those used in the centralized case. Specifically, 15 rounds were carried out at the server level and one epoch~ locally.

As in the centralized case, we performed cross-validation by dividing the dataset into two parts: training and testing in a ratio of $8$ to $2$, resulting in five models.

We performed two types of simulation.
The first one involved the use of almost the entire dataset, i.e.,~ $4000$ samples, increasing the number of clients, i.e., $4$, $8$, $16$ and $32$, collaborating in the training.
Thus, in~the four-client case, each client used a quarter of the samples used in the centralized case, i.e.,~ $1000$, while each doubling of the number of clients corresponded to a halving of the available samples.
As shown in Figure~\ref{fig:acc_combined}, in all the cases the accuracy is higher than $90\%$ also in the $32$-client case where each client is using only 125 samples as~a dataset.

The second one involved reducing the dataset available in the clients, i.e., from $1000$ representing the entire dataset to 50 samples representing one-twentieth of it, leaving the number of clients unchanged. 
As shown in Figure~\ref{fig:acc_combined}, we can obtain results above $90\%$ with only $125$ samples in each client in the case of a large number of clients, i.e.,~ $32$, and in all likelihood even with only 100 samples by increasing the number of clients. 

In addition,  increasing the dataset by a small amount, up to 250 samples, even just collaboration among a few clients, e.g., four, allows for accuracy rates comparable to those of the experienced pathologist, representing valuable technical support that can be used by physicians at the time of diagnosis. 

Instead, the use of a very small dataset, i.e., 50, does not allow for obtaining either acceptable or stable results, in fact, it turns out to be less than $60\%$ and deteriorates as the number of clients involved in the training increases. 

Another important result of this approach is shown in Table~\ref{tab:FL_statistic} and is that for~the same number of examples available in each client, increasing the number of clients involved in federated learning produces more stable and robust results.

\begin{table*}[ht!]
    \centering
    \begin{tabular}{c|c|c|c|c}
\hline
\textbf{\# Samples per Client} & \textbf{4 Clients} & \textbf{8 Clients} & \textbf{16 Clients} & \textbf{32 Clients}\\
\hline
            50 &57.67 ± 3.84 &57.61 ± 10.37 &0.5 ± 0.0 & \\
            100 &82.61 ± 4.30 &86.91 ± 3.33 &85.35 ± 3.91 &88.88 ± 3.87\\
            150 & & & &90.39 ± 0.63\\
            250 &91.28 ± 2.57 &92.21 ± 1.35 &93.24 ± 1.65 &\\
            500 &94.05 ± 1.17 & 93.63 ± 0.44& &\\
            750 &92.98 ± 1.79 & & &\\
            1000 &92.04 ± 0.99 & & &\\
\hline
\end{tabular}
    \caption{Accuracies (\%) of different architectures over the fivefold cross-validation setup: Avg. accuracy ± std. dev.}
    \label{tab:FL_statistic}
\end{table*}

\unskip

\section*{Discussion}\label{Discussion}

Although noninvasive systems for diagnosis and staging of NALFD are being studied and tested~\cite{reinshagen2023liver}, the visual analysis of histological sections on liver tissue samples obtained by biopsy or resection is still considered the gold standard. The histological analysis of steatosis is performed routinely to diagnose NAFLD or to decide about the suitability of liver grafts for transplantation. In addition, because NALFD can be associated with other diseases by increasing their risk~\cite{ko2023risk}, liver histological analysis plays a very important role in research and clinical settings.

There are previous works that used deep learning and NAFLD histology images, but these studies have taken into account animal tissues~\cite{heinemann2019deep, yang2019quantification}. Animal models of liver disease are required, but they cannot reproduce the complexity of the human liver. Tissue sample analysis from diagnostic biopsies can be more difficult compared to animal biopsies
performed during controlled experiments. Indeed, in~human tissue, there are many etiopathogenic factors (e.g., nutrition, drugs, viral infections) present in different combinations, which are difficult to quantify.

There are previous works using deep learning and histological imaging of NAFLD, but early studies considered animal tissues~\cite{heinemann2019deep, yang2019quantification} and more recent studies based on human tissues propose a steatosis segmentation model to identify steatosis individually~\cite{roy2020deep} or in~\cite{heinemann2022deep}. Heinemann~et~ al. perform the quantification of hepatic steatosis for histological evaluation of liver biopsies focusing mainly on the study of ballooning, inflammation, and the degree of separated fibrosis.

In our work we present a neural network that, in addition to being able to determine the transplantability of a liver with higher accuracy than an expert user, reduces the probability of false positives that would lead to transplanting a diseased liver into a healthy patient, potentially causing several complications.

Indeed, steatosis above the threshold value may lead to a higher incidence of initial poor graft function (IPGF), which has a negative prognostic impact as it is associated with an increased risk of renal failure, sepsis, and acute rejection in the first weeks after transplantation~\cite{arroyo2007advances,angeli2006switch}. Finally, some studies indicate that IPGF is associated with reduced organ and patient survival even in the long term~\cite{AISF}. 

Furthermore, by~taking advantage of the preprocessing tool that subdivides the initial image into smaller portions with the addition of a post-processing tool that reassembles these smaller images into the original, a computer-aided diagnosis system can be created to go alongside the medical device performing the examination and be used in real-time to determine the mapping of disease in a liver and help the physician to determine the potential development of the~disease faster.

In our hybrid quantum solution, we integrate quantum layers with traditional neural network architectures to exploit their ability to capture intricate correlations between data.
 Indeed, results show that such networks have more accurate classification predictions even on a small dataset as well as better generalization ability and less overfitting than their classical counterpart, which is a~ significant advantage in scenarios where large-scale data collection is challenging or impractical, such as in the medical field.
For these reasons, we believe that the HQNN can be considered for the image recognition task in the medical field.

Moreover, in Europe, discussions about rights in health care and privacy laws, such as the GDPR regulation, are of concern and are animating debates at the policy and legislative levels about the use of AI in the medical field.
Our study shows, and it is the first of its kind, that the use of federated learning complies with privacy laws without a significant loss of accuracy compared to a classical centralized approach.
Therefore, we believe that this technique can become the standard technique for creating computer-aided diagnosis systems that are robust, have high accuracy, and are designed while respecting privacy laws to support physicians in their daily work.

Our model \textit{seems} to exceed the results of an expert pathologist, but it still relies on data labeled by an expert pathologist, so it is currently still dependent on an external actor. To overcome this problem, the model can be updated in the future by relying on a new dataset labeled ``ground truth" that can be created by the user(s) after clinical and instrumental feedback on the correctness of the diagnosis and classification.

\subsection{Future~Works}\label{future_works}

In our federated model, we consider a fully balanced dataset among clients, where each client has the same number of samples equally distributed between transplantable and nontransplantable samples, which are always reachable. This situation is unlikely to be encountered by a model deployed in production. A valuable extension to this work would, therefore, be to investigate decentralized aggregation approaches with unbalanced datasets.

In addition, this study represents the core of a computer-aided diagnosis system currently under development, that automatically maps the presence of NAFLD in a liver by receiving as input its original image in jpeg2000 format and returning a ``heatmap'' image in which areas with different levels of disease have different coloring to provide the physician with an immediate mapping of the disease on the liver.

Another avenue to explore is the great potential of quantum technology to facilitate secure, distributed computation, as exemplified by the burgeoning frameworks for quantum federated learning \cite{Ren2023TowardsQFL}. 

\section*{Author contributions}\label{Author_contributions}
Lusnig and Cavalli designed and coordinated the study; Bonazza, Lusnig, Zanconati and Cavalli performed the classification, acquired the histological images and created the dataset; Sagingalieva, Surmach, Protasevich and Lusnig developed the classical solution; Sagingalieva, Melnikov, Surmach, and Protasevich developed the hybrid quantum solution; Lusnig, McLoughlin and Michiu developed the federated solution, De' Petris certified privacy solutions; all authors wrote the manuscript.

\bibliography{lib}

\end{document}